\documentclass[dvipsnames,format=sigconf,anonymous=false,review=false]{acmart}
\usepackage{listings}
\usepackage{xcolor}
\usepackage{graphicx}
\usepackage{subcaption}
\usepackage{float}

\AtBeginDocument{%
  }

\copyrightyear{2026}
\acmYear{2026}
\setcopyright{cc}
\setcctype{by-nc-nd}
\begin{document}

\title[Parallel Baymex for Clinical Data Classification]{Parallel Adaptive Multi-Objective Evolutionary Learning of Discretized Bayesian Network Classifiers for Clinical Data}

\author{Damy M.F. Ha}
\email{d.m.f.ha@lumc.nl}
\orcid{0000-0003-2495-9681}
\affiliation{%
  \institution{Leiden University Medical Center}
  \city{Leiden}
  \country{The Netherlands}
}

\author{Thalea Schlender}
\email{t.schlender@lumc.nl}
\affiliation{%
  \institution{Leiden University Medical Center}
  \city{Leiden}
  \country{The Netherlands}
}

\author{Yvette M. van der Linden}
\email{Y.M.van_der_Linden@lumc.nl}
\affiliation{%
  \institution{Leiden University Medical Center}
  \city{Leiden}
  \country{The Netherlands}
}

\author{Peter A.N. Bosman}
\email{Peter.Bosman@cwi.nl}
\affiliation{%
  \institution{Centrum Wiskunde \& Informatica}
  \city{Amsterdam}
  \country{The Netherlands}
}

\author{Tanja Alderliesten}
\email{T.Alderliesten@lumc.nl}
\affiliation{%
  \institution{Leiden University Medical Center}
  \city{Leiden}
  \country{The Netherlands}
}

\renewcommand{\shortauthors}{Ha et al.}

\begin{abstract} 
    Bayesian Networks (BNs) are of interest from an explainable AI viewpoint, offering transparent probabilistic models for decision support. 
    Baymex is a recently introduced multi-objective evolutionary algorithm for learning discretized BNs, enabling experts to trade-off different objectives of interest, such as likelihood, model complexity, and prior beliefs. 
    While Baymex has been shown to outperform state-of-the-art BN learning approaches, Baymex still 1) requires a lot of computation time and 2) has only been evaluated on synthetic data. To improve scalability, we introduce a parallelization strategy as well as a mechanism that enables adaptively steering optimization toward networks that overfit less. 
    We furthermore reconfigure Baymex to train a BN classifier through multi-objective optimization of cross‑entropy loss and the BIC complexity term so as to evaluate its performance on real-world clinical classification tasks. 
    Besides observing speedups up to over 54 times on a 16-core CPU, comparisons against clinically familiar baselines (decision trees, logistic regression, naive Bayes, and random forests) on two open-source (RADCURE and SUPPORT) and one in-house dataset, show that Baymex obtains statistically similar or better predictive performance while producing compact, clinically inspectable BNs. 
    Importantly, Baymex finds multiple plausible BN classifiers that contain predictors consistent with established clinical factors.
\end{abstract}

\ccsdesc[300]{Computing methodologies~Parallel computing methodologies}
\ccsdesc[500]{Computing methodologies~Bayesian network models}
\ccsdesc[300]{Computing methodologies~Randomized search}
\ccsdesc[100]{Computing methodologies~Machine learning}

\keywords{Bayesian networks, evolutionary algorithms, multi-objective optimization, parallelization, adaptive steering, explainable AI}

\received{20 February 2007}
\received[revised]{12 March 2009}
\received[accepted]{5 June 2009}

\maketitle

\section{Introduction}
\label{s:introduction}
Every day, clinicians integrate symptoms, comorbidities, imaging, and lab results to form probabilistic judgments about diagnosis, treatment response, and
survival in order to guide their patients. Providing accurate data-driven decision support tools can help improve
the consistency, transparency, and overall quality of these decisions. Although
modern AI methods have demonstrated substantial promise across a wide range of
clinical tasks, their adoption in practice remains limited, partly because most
models are not sufficiently transparent~\cite{kyrimibn_review_2021}.

Bayesian Networks (BNs) offer a trusted and tried alternative rooted in classic
AI. BNs provide joint distributions over variables using a Directed Acyclic
Graph (DAG) and conditional probability tables, allowing explicit modeling of
conditional structure and enabling transparent probabilistic
reasoning~\cite{book_pearl_1988, book_koller_friedman_2009}. In contrast to many
black-box predictors, both the learned structure (the DAG) and the conditional
probability tables can be inspected and challenged by domain experts. This makes
BNs particularly interesting from an eXplainable AI (XAI) perspective
\cite{example_cardiovascular_breast_cancer, example_atrial_fibrillation_relapse,
kyrimibn_review_2021}.

Despite these benefits, learning even the structure (i.e., the DAG) of a BN from
data remains computationally challenging \cite{learning_bn_npcomplete,
kitson_survey_bn_structure_learning_2023}. As data collection continues to grow
in both the number of samples and features, performing (optimal) structure learning
becomes progressively more intractable. Moreover, in many clinical applications,
the data contains real-valued features, which are usually discretized in
order to build a BN (e.g., ~\cite{example_discretization_prostate_cancer, example_discretization_lung_cancer, example_discretization_rectal_cancer}). However, the discretization procedure, including the number
of bins used, shapes the BN structure that is recovered by a structure learning
algorithm. Different binning choices modify the conditional probability tables
in distinct ways, thereby influencing which dependencies get selected.

The recently introduced \textbf{Baymex} (\textbf{Bay}esian network learning via
evolutionary intelligence for \textbf{m}ulti-objective \textbf{ex}plainable AI)
algorithm \cite{baymex} addresses part of this challenge by learning discretized
BNs using a combination of state-of-the-art multi-objective Evolutionary
Algorithms (EAs)~\cite{mo_gomea, gi_gomea, mo_rv_gomea}. This results in a
search for solutions that best trade off objectives of interest, such as
likelihood, model complexity, and agreement with expert beliefs. Accordingly,
instead of providing a single "best" network, Baymex produces an approximation
of the Pareto set, containing multiple plausible BNs that trade-off the
objectives of interest. Besides being able to glean additional insights from
inspecting the optimized trade-offs, practitioners benefit from the option to
choose the most appropriate model~\cite{baymex}. This offers greater flexibility
and chances of uptake for real-world use compared to having to accept a single
model returned by the vast majority of BN learning
methods~\cite{example_atrial_fibrillation_relapse,
example_bn_mental_health_disorder, example_neurocognitive_disorder,
example_expert_bn}. While Baymex has been shown to outperform state-of-the-art
BN learning approaches, Baymex still 1) requires a lot of computation time and
2) has only been evaluated on synthetic data. In this work, we address both
limitations.

Firstly, EAs are naturally amenable to parallelization because they usually
evaluate many candidate solutions per generation independently. Consequently,
parallel evolutionary computation is supported by a substantial and
well‑established body of literature~\cite{parallel_ea, evo_x_parallel,
arthur_parallel_2023}. In this work, we parallelize Baymex and investigate a mechanism that enables adaptively steering
optimization to focus as much as possible only on networks that do not overfit.

Secondly, we reconfigure Baymex to train a BN classifier through multi-objective
optimization of cross-entropy loss and the BIC complexity term~\cite{bic}. By
doing so, we are able to tackle real-world classification tasks, finding BN classifiers
that are optimized for trading off predictive performance and model complexity.

We evaluate the accelerated and fundamentally improved Baymex
algorithm\footnote{The code for Baymex is
available at: \url{https://github.com/damyha/baymex}.
} on real-world clinical classification tasks. To this end, we include the
SUPPORT dataset, comprising a large cohort originally designed for prognostic
modeling of 180-day survival in seriously ill hospitalized
adults~\cite{support_original, Knottenbelt2025CoxKAN}. We also include the
RADCURE dataset, comprising an open-access head-and-neck radiotherapy cohort for
which we learn a classifier to predict whether patients achieve 2‑year survival~\cite{radcure}. Finally, we revisit
a study of patients with advanced cancer led by our radiotherapy department on
survival in patients with spinal bone metastases and use Baymex to predict, via classification, a
minimum 3-month survival~\cite{dataset_spinal}. We furthermore make comparisons
with widely-used, clinically familiar baselines: logistic regression, decision
trees, naive Bayes, and random forest.

Besides observing speedups up to over 54 times on a 16-core CPU, across datasets, we find that Baymex obtains statistically similar or better predictive performance than the explainable baselines and, in some cases, better than all baselines while producing transparently inspectable DAGs
and conditional probability tables. Importantly, Baymex finds multiple plausible
BN classifiers that contain predictors consistent with established clinical
factors.

\section{Background}

\subsection{Bayesian Networks}
\label{ss:bayesian_networks}
A Bayesian Network (BN) is a probabilistic graphical model that represents a factorized joint distribution over a set of random variables~\cite{book_pearl_1988}.
A BN is defined by a Directed Acyclic Graph (DAG) whose vertices represent random variables and whose directed edges (i.e., arcs) specify parent–child relationships used to factorize the joint distribution into local conditional probability distributions.
Specifically, let $\mathbf{X} = (X_1,\dots,X_N)$ denote $N$ random variables and let $G=(V,E)$ be a DAG whose vertices $V$ correspond to the random variables and edges $E$ connect the vertices.
A BN factorizes the joint probability distribution as follows: $p(\mathbf{X}) = \prod_{i=1}^{N} p\!\left(X_i \mid \mathrm{pa}(X_i)\right)$,
where $\mathrm{pa}(X_i)$ denotes all the variables corresponding to the parent
vertices of the $i$-th vertex in the DAG.

\subsection{Baymex: Learning Discretized BNs}
\label{ss:baymex}
To understand the proposed improvements that we introduce to Baymex in this
paper, we will list core concepts of Baymex here, but we refer the interested
reader to the literature for more details~\cite{baymex}.

\subsubsection{Mixed Discrete/Real-valued Decision Space}
A central feature of Baymex is its capability to perform multi-objective
optimization in mixed discrete and real-valued search spaces. Baymex models
a BN's DAG structure using a discrete encoding scheme. For a network of
$N$ vertices, the space of potential edges is defined by $\ell_{\text{total}} =
\frac{N(N-1)}{2}$ variables. Each potential edge between vertex $i$ and $j$
($j>i$ and $i,j \leq N$) is modeled as a ternary decision variable $\in \left\{
0, 1, 2\right\}$, representing no connection, a directed edge $i \rightarrow j$,
or a directed edge $j \rightarrow i$, respectively. To ensure that every solution
encodes an acyclic graph, Baymex implements a repair mechanism that identifies
and repairs cyclic dependencies.

\subsubsection{Discretization.}
To handle real-valued data, Baymex discretizes real-valued data during search
using bins that are encoded into the solution. Specifically, for each
real-valued feature $j$, a bin count $B_j$, i.e., the number of bins used to
discretize the real-valued data, is encoded as a discrete decision variable
between a minimum of 2 and a user-defined maximum $B_j^{\text{max}}$. In Baymex,
bin boundaries are defined as the midpoints between consecutive samples. To this
end, the samples are sorted and duplicate values are deleted. An index in this
sorted list then describes between which two consecutive samples the bin
boundary is defined. As there are typically many indices, a real-valued decision
variable is used to this end, which is rounded to the nearest integer for
evaluation. Because depending on the number of bin boundaries to use, the best
bin boundaries may be in very different locations, in Baymex $n-1$ distinct
indices are used for $n$ bins for all $n \in \{2,3,\ldots,B_j^{\text{max}}\}$:
$$
\left [
  \begin{array}{c}
  \!
  \underbrace{b_{21}^j}_{\text{2 bins}},
  \underbrace{b_{31}^j, b_{32}^j}_{\text{3 bins}},
  \underbrace{b_{41}^j, b_{42}^j, b_{43}^j}_{\text{4 bins}},
  \cdots
  \!
  \end{array}
\right ]
$$
Here, $b_{ki}^j$ is the $i$-th boundary used to separate variable $j$ into $k$
bins. The value of $B_j$ determines which of the bin boundaries in the
representation are active. The other bin boundaries remain in the solution, but
are inactive, i.e., they are introns.

\subsubsection{Variation via Gene-pool Optimal Mixing}
Baymex combines core components drawn from different variants of the Gene-pool
Optimal Mixing Evolutionary Algorithm (GOMEA), adapting them to the requirements
of BN learning. For the specific goal of multi-objective optimization, Baymex draws upon a key component of multi-objective GOMEA~\cite{mo_gomea}: clustering. Each generation, the population is clustered into k roughly equal groups in objective space, each addressing part of the approximation front.
For each cluster, Baymex configures variation models separately, to tailor variation
to the local structure of each
region.  Baymex makes use of three variation operators:
GOM~\cite{mo_gomea}, GI-GOM~\cite{gi_gomea}, and RV-GOM~\cite{mo_rv_gomea}, targeting BN structure, bin counts, and bin boundaries, respectively. Each of
these operators combine variation and selection by
sampling new values for a subset of all variables in a solution and accepting the change only if it does not worsen the solution or if the solution can be added to the elitist archive. Otherwise, the
changes are reverted. Each such subset is called a linkage set. How the linkage
sets and the sampling procedure are configured, differs per type of variable, for which we kindly refer to
the literature~\cite{baymex, mo_gomea, gi_gomea, mo_rv_gomea}.

\subsubsection{Elitist Archive}
To retain the non-dominated solutions found during the search, Baymex maintains an elitist archive. While the implementation supports capacity constraints
(see~\cite{hoang_elitist_archive}), in our experiments, the number of
non-dominated solutions remains relatively small compared to the target archive
size (10,000 solutions). As a result, the archive functions as a regular elitist
archive.

\subsubsection{Algorithmic Structure}
Algorithm~\ref{lst:loop_baymex} summarizes the key aspects of a Baymex generation and explicitly shows the nested loops over clusters, linkage sets, and solutions. Additionally,
it is highlighted where offspring are tested for entry into the elitist
archive. For discrete variables, this is immediately after variation of a solution.
For real-valued variables, this is after every solution in a cluster has undergone variation.

\begin{minipage}{80mm}
\renewcommand{\lstlistingname}{Algorithm}
\begin{lstlisting}[basicstyle=\ttfamily\fontsize{7.5}{9}\selectfont,label={lst:loop_baymex},caption={Pseudocode of Baymex's optimization loop.}]
all_offspring_clusters = []
for each cluster
  cluster_offspring = []
  for each solution in cluster:
    cluster_offspring.append(clone(solution))
  for each linkage_set in cluster
    if linkage_set.type == DISCRETE_GOMEA:
      for each solution in cluster_offspring:
        apply_GOM(solution, linkage_set) # Parallelizable
        add_to_archive(solution)
    elif linkage_set.type == DISCRETE_GI-GOMEA:
      for each solution in cluster_offspring:
        apply_GI-GOM(solution, linkage_set) # Parallelizable
        add_to_archive(solution)
    elif linkage_set.type == REAL-VALUED_GOMEA:
      offspring_for_archive = []
      for each solution in cluster_offspring:
        apply_RV-GOM(solution, linkage_set) # Parallelizable
        offspring_cluster.append(solution)
      for each solution in offspring_for_archive:
        add_to_archive(solution)
  all_offspring_clusters.append(cluster_offspring)
replace_population_with_offspring(all_offspring_clusters)
\end{lstlisting}
\end{minipage}

\subsubsection{Interleaved Multi-start Scheme}
In Baymex, the need for manual tuning of the population size and number of clusters is avoided through the use of an Interleaved Multi‑start Scheme (IMS).
With IMS, several Baymex instances are managed, each with a distinct population size
and number of clusters. The IMS uses a base population size of 30, and a base
number of clusters of $k = m + 1$, where $m$ is the number of optimization
objectives. Every 8 generations of a Baymex instance, one generation of a Baymex instance is performed with twice the population size and
one extra cluster. For more details of Baymex we kindly refer to \cite{baymex}.

\subsection{Bayesian Network Classifier}
\label{ss:bayesian_network_classifiers}
BNs can be turned into probabilistic classifiers by designating one random
variable as the class label $Y$ and modeling the joint distribution
$p(Y,\mathbf{X})$, from which predictions follow via the posterior $p(Y \mid
\mathbf{X})$. BN classifiers have been studied for
decades~\cite{bn_classifiers}, spanning a spectrum from simple naive Bayes,
which assumes feature independence given the class, to more flexible
BN structures that can capture more complex dependencies~\cite{
tree_augmented_naive_bayes, bn_classifiers_2014}.

\section{Methodology}

\subsection{Baymex Classifier}
In this work, we use BN classification as a real-world test-bed for Baymex.
Accordingly, we adapt Baymex from learning BNs to learning BN classifiers by
minimizing the training cross-entropy loss. 
Conditional probabilities are estimated using Dirichlet smoothing with a uniform Dirichlet prior to handle samples unseen during training.
We retain the core feature of multi-objective optimization in Baymex and use the BIC complexity term as a second minimization objective to trade-off accuracy and model complexity. 
For BNs, this is $\sum_{i=1}^{N}
(B_i-1)\prod_{X_j\in pa(X_i)}B_j\cdot \log\left(\frac{n}{2}\right)$, where $n$
is the sample size.
We keep the search space unchanged so that improvements obtained with Baymex for classification remain directly translatable to BN learning.

\subsection{Parallelization}
\label{ss:parallelization}
As can be seen in Algorithm~\ref{lst:loop_baymex}, variation in Baymex proceeds by by looping over clusters and, within each cluster, over linkage sets, applying variation to every solution. Consequently, parallelization of solution variation and evaluation can be done straightforwardly for the innermost loops (Lines 8, 12, and 17). 
However, updating the elitist archive remains a strictly sequential operation, necessitating the use of mutex semaphores (i.e., locks that ensure that only one parallel thread can access the elitist archive at a time).

Furthermore, the GI-GOM operator cannot be parallelized without making changes on an algorithmic level because in GI-GOM every solution exchanges genes with a random other solution sequentially. 
Therefore, concurrent threads can interfere by selecting the same partners, violating the operator’s sequential design. 
For parallelization, alternative formulations are required which pre-compute exchange partners. 
Although this does not diminish the speed‑ups achievable through parallelization, it introduces minor algorithmic differences relative to the original GI‑GOM. As demonstrated in the supplementary materials, these differences do not affect optimization performance compared to the original GI‑GOMEA~\cite{gi_gomea}.

Finally, Parallelization of GOM has no algorithmic implications, as each variation can be done fully independently.

\subsection{Adaptive Steering}
\label{ss:adaptive_steering}
\begin{figure}
    \centering
    \vspace*{1mm}
    \includegraphics[width=0.9\linewidth]{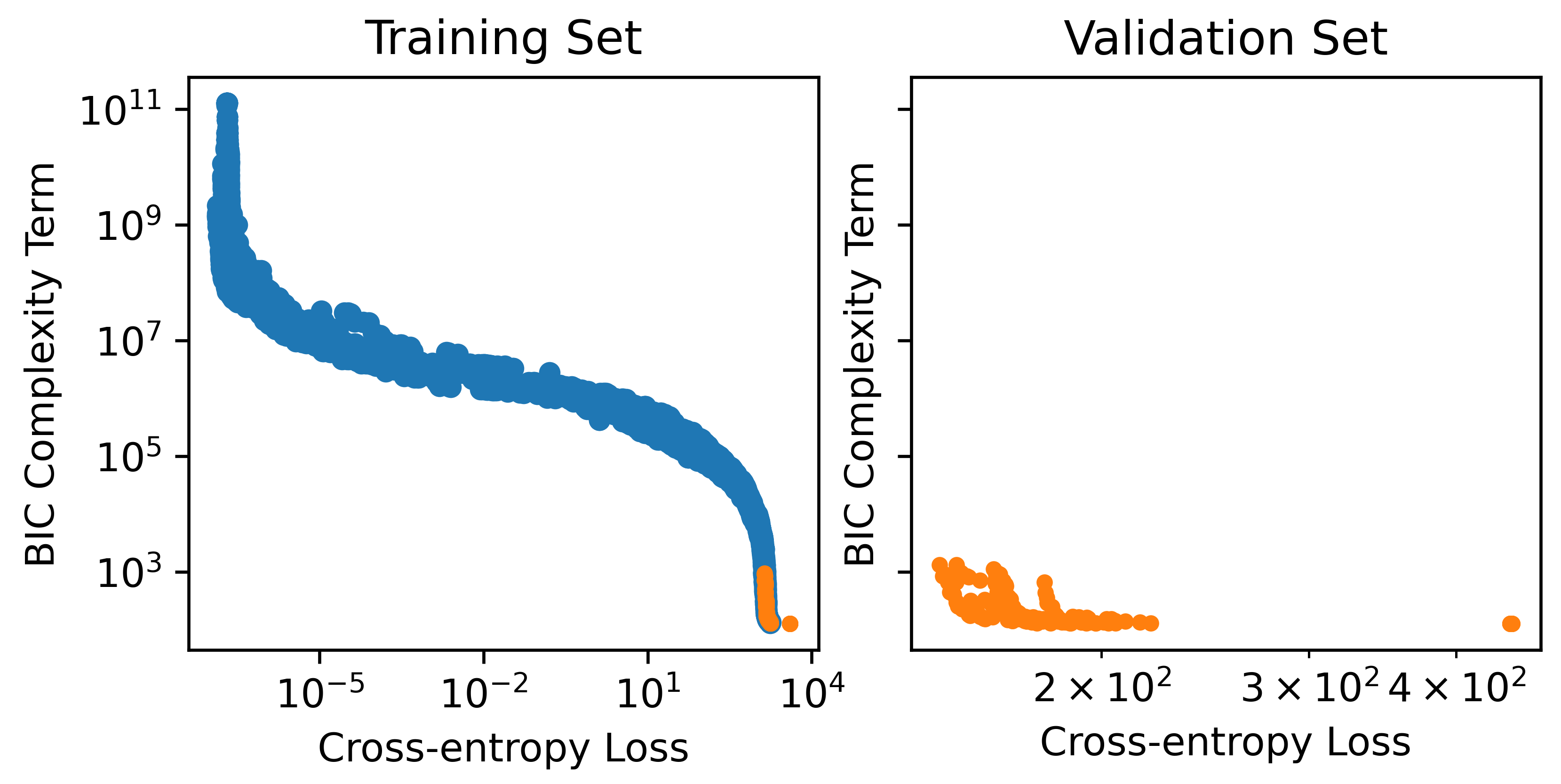}
    \vspace*{-4mm}
    \caption{All approximation fronts (cross-entropy loss vs complexity) over 30 train-validation splits for the training set (left) and validation set (right). Corresponding solutions in the training and validation fronts are shown in orange.}
    \Description{ }
    \label{fig:example_front}
    \vspace*{-8mm}
\end{figure}

Early experiments with Baymex for learning BN classifiers resulted in large
approximation fronts on the training data. However, when evaluated on validation
data, many BNs became dominated, indicating substantial overfitting.
Figure~\ref{fig:example_front} illustrates this behavior on the SUPPORT dataset
(see more in Section~\ref{ss:datasets}) for which we performed 30 independent
runs on 30 different train-validation splits.

Figure~\ref{fig:example_front} (left) shows the approximation fronts obtained on the training folds, with training cross-entropy loss plotted against the BIC complexity term. 
The resulting BN classifiers cover a wide complexity range, going up to approximately $10^{11}$. 
We then reevaluate the cross-entropy loss and BIC complexity term of these same BNs on the corresponding validation sets and retain only the non-dominated solutions. These validation fronts are shown on the right in Figure~\ref{fig:example_front}. The resulting set of BNs constitute a much more narrow range in terms of complexity, going up to only $10^{3}$.

Baymex appears to spend a substantial part of its computation budget exploring highly complex BN classifiers that perform well on the training set, but are costly to evaluate. Yet, many of these models become Pareto‑dominated once validation cross‑entropy is considered, indicating that they do not provide real accuracy–complexity benefits and are likely overfitting.
This observation motivates the use of mechanisms that bias the search towards regions that are more likely to generalize well, e.g.,~\cite{desirable_objective_ranges,
desired_pareto_front_with_ref_point,adaptive_steering}.

In this work, we take inspiration from~\cite{adaptive_steering} and augment Baymex with a complexity cap that is adaptively guided by observations on the
validation data set during optimization. 
In addition to having an elitist archive based on training-data objective values, we maintain a second elitist
archive based on validation-data objective values. 
At the start of each generation, the current most complex BN is extracted from the validation elitist archive. 
The complexity of this solution is used as an adaptive reference threshold in the form of a hard constraint, denoted ${\mathrm {BIC}}_\theta$. 
Each solution $x$ that gets evaluated in this generation, is assigned a constraint value that represents the excessive complexity of that solution, relative to this
threshold: $\text{C}(x) = \max\{0, {\mathrm {BIC}}(x) - f_c{\mathrm
{BIC}}_\theta\}$, where ${\mathrm {BIC}}(\cdot)$ denotes the BIC complexity term and $f_c$ determines the leeway we allow to discover BNs that are more complex, but have better accuracy. 
Solutions with BIC complexities smaller than $f_c{\mathrm {BIC}}_\theta$ remain unconstrained, while solutions exceeding $f_c{\mathrm {BIC}}_\theta$ incur increasing constraint values. 
We subsequently use constraint domination~\cite{deb2000efficient} to ensure that any solution that becomes infeasible during variation is rejected, and any
solution that is already infeasible is only modified in ways that reduce its constraint violation. 
Empirical evaluation showed that a leeway factor of $f_c = 10$ provides sufficient flexibility for Baymex to discover more complex and accurate solutions, ensuring that the adaptive constraint does not overly restrict the search.

Since the constraint reference threshold is potentially changed each generation, solutions that were previously constrained may be no longer constrained in a new
generation, or vice versa. 
For this reason, after the constraint reference threshold has been set, we recompute the constraint values of all solutions in the population. 
Additionally, we empty the training elitist archive and reinsert those original solutions that satisfy the complexity constraint.

\section{Experimental Setup}
\subsection{Training Pipeline}
\label{ss:baymex_training_pipeline}
We compare Baymex against well-established baselines using a common outer train-test split.
For the baselines (see Section~\ref{ss:baseline_classification}), model selection is performed by standard 30-fold cross-validation on the training set to obtain many validation estimates of Baymex. 
Within each fold, candidate models are trained on the training subset, evaluated on the corresponding validation subset, and the model with the best validation score is selected.
The final performance is then reported on the held-out test set.

\begin{figure}[htbp!]
    \vspace*{-2mm}
    \centering
    \includegraphics[width=1\linewidth]{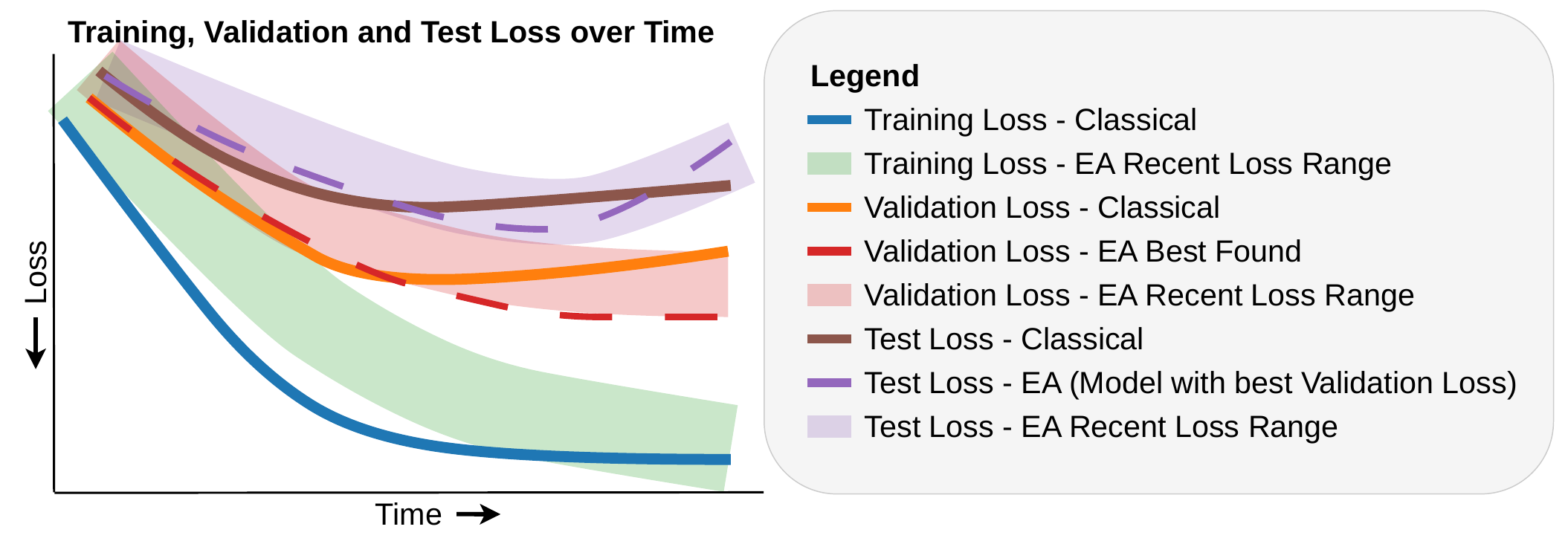}
    \vspace*{-7mm}
    \caption{Example of validation‑loss overfitting with an EA compared to a gradient‑based method. EA search evaluates many more solutions, some with worse training loss but better validation loss, which increases the risk of overfitting when selecting the model with the best validation score.}
    \label{fig:example_train_vs_validation_loss}
    \Description{ }
    \vspace*{-2mm}
\end{figure}

Baymex can follow the standard model‑selection procedure, but it can also use an alternative approach in which a refinement set is withheld from the training data to guide selection from the validation archive. This is motivated by the EA‑based optimization underlying Baymex: unlike gradient‑based methods, it is not restricted to a greedy improvement path in training loss. EA‑based search explores a wider range of loss values, and while some solutions may degrade on the validation set, others may achieve validation scores that surpass the previous best (see Figure~\ref{fig:example_train_vs_validation_loss}).
Although this stronger exploratory capability improves optimization, it also increases the risk of overfitting when selecting solely based on validation performance. The refinement set therefore provides an additional selection step that mitigates this risk. Because overfitting is typically less severe on the validation set than on the training set, using a separate refinement set is expected to yield the best results when sufficient data is available. However, when data is scarce, splitting off a refinement set may itself induce overfitting, where models are selected by validation performance from the final elitist archive based on training loss, preferable.

\subsection{Datasets}
\label{ss:datasets}
We evaluate Baymex and the baselines on three clinical datasets (described below) with a binary overall‑survival endpoint, resulting in a binary classification task. 
When a dataset does not define a fixed-horizon target, we set a clinically meaningful prediction horizon and convert the recorded overall-survival time into a horizon-based label where $1$ means survival, and $0$ means the patient has deceased before the prediction horizon. Since most datasets include censored individuals (i.e., the event has not yet occurred at analysis time), the true survival time is not always observed. 
We therefore consider each patient's minimum survival time for the classification task, defined as the last available
follow-up time (time of death if observed, censoring time otherwise).

\paragraph{SUPPORT}
This is a cohort of hospitalized adults. 
The data contains 14 features including demographic variables, comorbidities, chronic comorbidities, physiological measurements, and laboratory values~\cite{support_original}. 
We use a publicly available preprocessed version of this data~\cite{Knottenbelt2025CoxKAN} and create a training/validation ($N_{\text{train}}=7,985$) and held-out test ($N_{\text{test}}=888$) partition that has no missing values. 
We applied semantic remapping (real-valued to ordinal or categorical) where clinically meaningful. 
The exact remapping is provided in our supplementary material.


\paragraph{RADCURE}
This is an open-source head-and-neck cancer radiotherapy dataset comprising tabular features and CT images~\cite{radcure}. 
Here, we only use tabular features. 
We removed patients with missing values and collapsed rare
discrete values based on methods in \cite{malafaia2025automatedinterpretablesurvivalanalysis}. 
We used the provided time-to-event feature to predict two-year survival, as was done in the RADCURE
challenge~\cite{radcure}.
The selected features (16) and preprocessing steps (filtering and remapping) are fully specified in our supplementary material. 
After preprocessing, the cohort is reduced to a training/validation set of $N_{\text{train}}=2,154$ samples and a held-out test set of $N_{\text{test}}=364$ samples.

\paragraph{Spinal Metastases}
This is a dataset from a study led by our clinical department on the survival of patients with advanced symptomatic spinal bone metastases who underwent radiotherapy and/or surgery~\cite{dataset_spinal}. 
The training/validation set contains $N_{\text{train}}=1,043$ patients from a single-centre retrospective cohort of patients treated between January 2001 and December 2010. 
As an external validation set, 339 patients that had
symptomatic spinal bone metastases are taken from the national Dutch bone metastases study~\cite{DBMS}. 
The datasets contain 16 features and are preprocessed according to \cite{dataset_spinal, pisa} and a minimum three-month survival is predicted. 
Due to privacy regulations, this dataset is not publicly available.

\subsection{Baymex Efficiency Enhancements}
All runs to study Baymex enhancements were executed on an AMD EPYC Genoa 9654 with a wall‑clock budget of 24 hours per fold. Here, we use the SUPPORT dataset, as it is our largest dataset and thus the case where search‑efficiency enhancements are most needed.
Because larger starting populations in IMS typically require more time to reach good solutions, whereas smaller populations offer less parallelization potential, we considered a range of mimimum IMS population sizes: 30, 120, 480, and 1920. Three types of enhancements were studied.
 
\subsubsection{Parallelization}
We ran Baymex in two configurations: the original serial version on a single CPU core, and the parallelized version on 16 CPU cores. 

\subsubsection{End-Of-Generation (EOG) Elitist Archive Update}
We study how updating the elitist archive at the end of each generation, rather than during variation, affects the search efficiency of Baymex.

\subsubsection{Adaptive Steering}
We study how adaptive steering influences the search efficiency of Baymex. 

\subsection{Clinical Data Classification}
All runs of Baymex for clinical data classification were done with the parallel variant of Baymex with adaptive steering. The experiments on the SUPPORT and RADCURE datasets were performed on 16 AMD EPYC Genoa 9654 cores.
Due to privacy restrictions, experiments on the spinal metastases dataset were performed on our
in-house HPC system using 16 Intel E-2690 cores.

\subsubsection{Baseline Classification Algorithms}
\label{ss:baseline_classification}
We compare Baymex to four widely-used, well-established baselines.

\begin{enumerate}
    \item Decision Tree (DT)
    \item Random Forest (RF)
    \item Binary Logistic Regression (LR)
    \item Naive Bayes (NB)
\end{enumerate}

We explicitly do not include models based on deep neural networks as they are
not inherently interpretable. While the same can be said for RF, we include it
because despite its limited intrinsic interpretability, it is often used as
a standard baseline for tabular data, known for its strong and
often‑reported accuracy.

We use the same 30 training splits for all algorithms. For
each baseline, we evaluate various hyperparameter settings (see 
supplementary material) and report the results obtained with the settings
that achieved the best validation or refinement performance.

\paragraph{Decision Tree}
A DT classifies data by repeatedly splitting data along branches based on
features that best separate the target classes. A single DT can be highly
interpretable when it remains small. Furthermore, a smaller tree can reduce
overfitting. Therefore, for DT, we tune the maximum depth, minimum samples per
split, and cost-complexity pruning strength (ccp\_alpha).

\paragraph{Random Forest}
An RF is an ensemble model that combines the predictions of multiple
DTs, each independently trained on random subsets of samples and features. RF
provides a strong non-linear baseline, at the cost of reduced interpretability
relative to a single DT. We train an RF classifier on the cleaned dataset and
tune the number of trees, maximum depth, number of features to select per tree,
minimum number of samples per split, and pruning strength.

\paragraph{Binary Logistic Regression}
Binary LR is a statistical method that models the probability of a two-category
outcome by linking features to the log-odds of the outcome occurring. LR is
considered interpretable because its coefficients correspond directly to
log‑odds changes. Some data preprocessing is however required. Real-valued data
is standardized using z-normalization, categorical data is one-hot encoded, and
ordinal data is left as is. We tune the solver, regularization type,
regularization strength, and for the elastic-net solver the $\ell_1$ mixing
parameter.

\paragraph{Naive Bayes}
NB is a probabilistic baseline that assumes conditional independence
of features given the outcome. It can be viewed as a simplified analogue of BN
classifiers, without additional learned structure. 
To preserve interpretability, we restrict NB to discrete features.
For this, we either remove real-valued features or pre-discretize the variables using the technique of \cite{FayyadIraniDiscretization}. We also tune a univariate
feature‑selection step (mutual information or ANOVA $F$-score ranking) and the
Laplace smoothing strength.

\subsubsection{Impression of Baymex's Multi-Objective Front}
\label{ss:objective_space_clustering}
To characterize the quality of the multi-objective approximation front obtained by Baymex, we aggregate all solutions over all folds and partition them into five complexity-based bins to get an idea of test performance in the groups of models of extreme low- and high-complexity, the knee region, and the intermediate regions between them.
We construct the bins using both Equal-Width (EW) and Equal-Frequency (EF) discretization.
For each bin, we report the distribution of test ROC-AUC, providing a clearer picture of how predictive performance is spread across the obtained front.

\subsubsection{Statistical Testing}
\label{ss:statistical_testing}
For each prediction task, we compare Baymex to the baseline algorithms described
in Section~\ref{ss:baseline_classification} using the held‑out test set. To do so,
one needs to perform final model selection, however. Here, for simplicity, we choose
the best model on the validation or refinement set. However, an expert may in practice
choose to study the models and choose a different trade-off between complexity and
validation (or refinement) accuracy. Importantly, the model with the best validation
score may not be the model with the best test score. This could be
another model that could be selected by an expert. For this reason we show
test scores along the entire front, in bins, in Figure~\ref{fig:cluster_distribution}.
For the best validation/refinement scoring model, we
report the median, and 25\%-75\% quantiles, test ROC‑AUC per algorithm and
identify the best‑performing algorithm. We then assess whether this best
algorithm performs significantly better than each competing algorithm using a
paired Wilcoxon signed‑rank test with $\alpha=0.05$. To correct for multiple
testing, we apply a Bonferroni correction. In total, we conduct 34 tests,
yielding a corrected significance level of $\alpha_{\mathrm{corr}}=\alpha/34$.

\section{Results}
\subsection{Baymex Efficiency Enhancements}
\subsubsection{Parallelization}
\label{ss:baymex_parallelization}
\begin{figure}[htbp!]
    \vspace*{-4mm}
    \centering
    \includegraphics[width=1\linewidth]{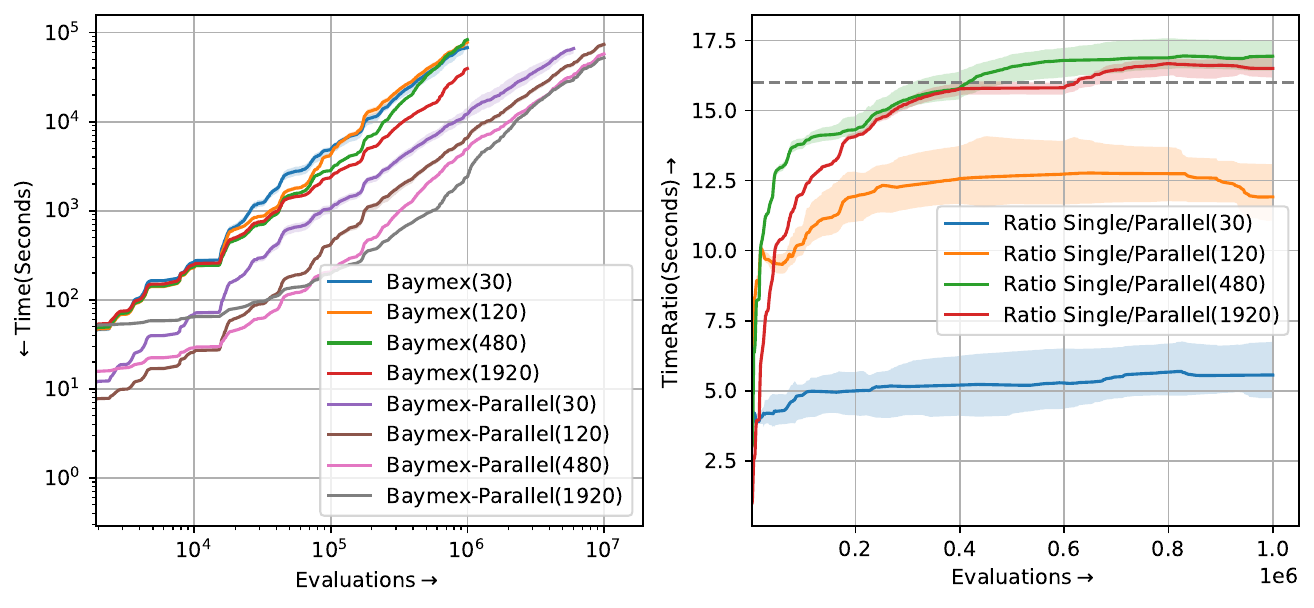}
    \vspace*{-8mm}
    \caption{Evaluation throughput of serial and parallel Baymex on SUPPORT across minimum IMS population sizes.
    Left: wall‑clock time at matched evaluation counts. 
    Right: serial-to-parallel wall-clock ratios.}
    \Description{ }
    \vspace*{-1mm}
    \label{fig:baymex_parallel_evaluations}
\end{figure}

As expected, parallelization yields substantial speedups, increasing with larger IMS minimum population size, as can be seen in Figure~\ref{fig:baymex_parallel_evaluations}. 
Interestingly, for IMS starting population sizes 480 and 1920, the time ratio exceeds the upper bound of achievable efficiency. The reason for this is that BNs with higher complexity take longer to evaluate. In parallel Baymex, BNs of varying complexity are generated at different stages of the search, causing solutions to enter the elitist archive in a different order than in the single‑core setting and resulting in different convergence behavior and solution types being evaluated.

\subsubsection{End-Of-Generation (EOG) Elitist Archive Update}
\label{ss:baymex_eog_update}
\begin{figure}[htbp!]
    \centering
    \vspace*{-3mm}
    \includegraphics[width=1\linewidth]{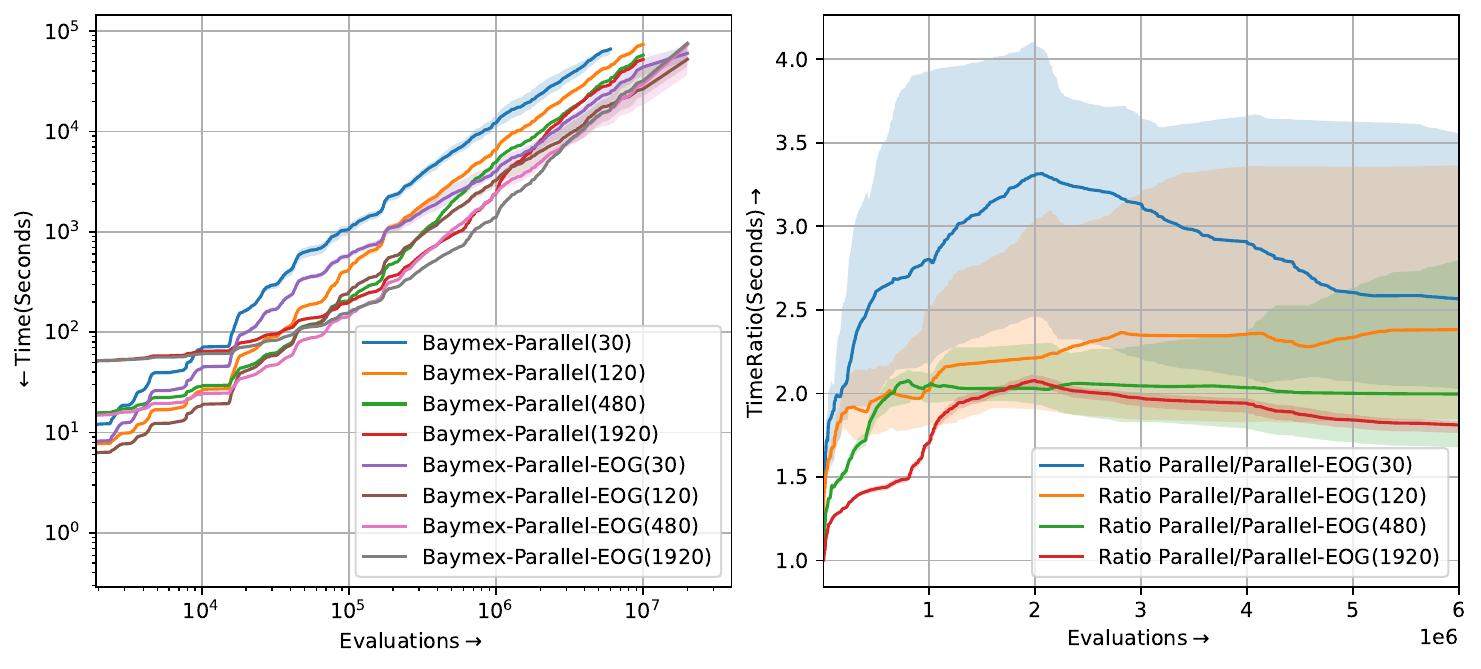}
    \vspace*{-8mm}
    \caption{Impact of EOG elitist‑archive updates in Baymex‑Parallel on SUPPORT, evaluated across IMS minimum population sizes. Left: wall‑clock time at matched evaluation counts. Right: wall‑clock time ratio.}
    \Description{ }
    \label{fig:exp_end_of_generation}
\end{figure}

Frequent access to the mutex‑protected elitist archive can become a bottleneck. Pushing all elitist archive operations to the end of a generation potentially alleviates this.
As expected, reducing the frequency of elitist archive updates yields additional speedups, particularly for smaller minimum population sizes in IMS, as can be seen in Figure~\ref{fig:exp_end_of_generation}.

\subsubsection{Adaptive Steering}
\label{ss:exp_adaptive_steering}

\begin{figure}[htbp!]
    \vspace*{-4mm}
    \centering
    \includegraphics[width=1\linewidth]{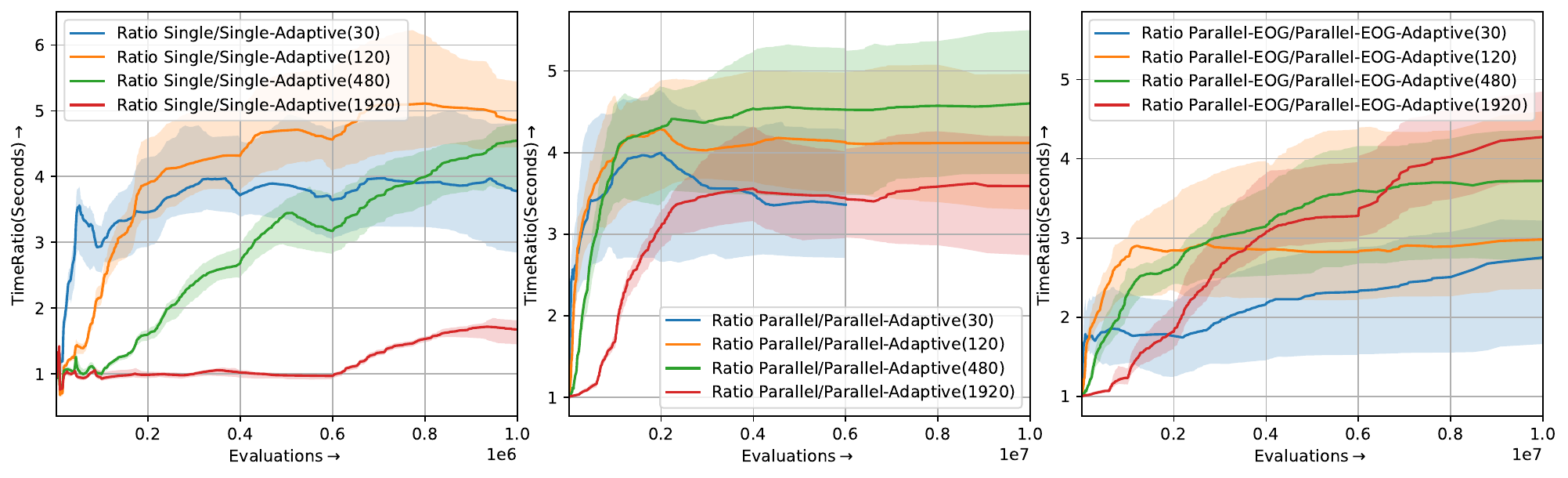}
    \vspace*{-7mm}
    \caption{Effect of adaptive steering on the SUPPORT wall-clock time ratio for serial and parallel Baymex, with and without EOG updates, across minimum IMS population sizes.}
    \Description{ }
    \label{fig:parallel_evaluations_adaptive}
\end{figure}

Figure~\ref{fig:parallel_evaluations_adaptive} shows the time ratios of Baymex, Baymex-Parallel, and Baymex-Parallel-EOG relative to their adaptive variants.
With the exception of Baymex-Adaptive(1920), adaptive steering yields additional speedups. This is because adaptive steering prevents the construction of overly complex networks, which are substantially more time‑consuming to evaluate.
Interestingly, Baymex‑Adaptive(1920) shows no speedup before 600,000 evaluations and is even slower than the non‑adaptive variant before 100,000. This stems from the warm‑up required for the adaptive reference point, combined with the extra evaluations needed to re‑evaluate constraint values for the entire population.

Overall, these results, together with those from the previous subsection on parallelization, show
that to fully exploit $N$ parallel computation cores, it is advantageous to
start from an IMS population size that is (much) larger than the default 30 for
serial Baymex.

\subsubsection{Impact of Baymex Enhancements on Accuracy}
\begin{figure}[htbp!]
    \centering
    \vspace*{-3mm}
    \includegraphics[width=0.8\linewidth]{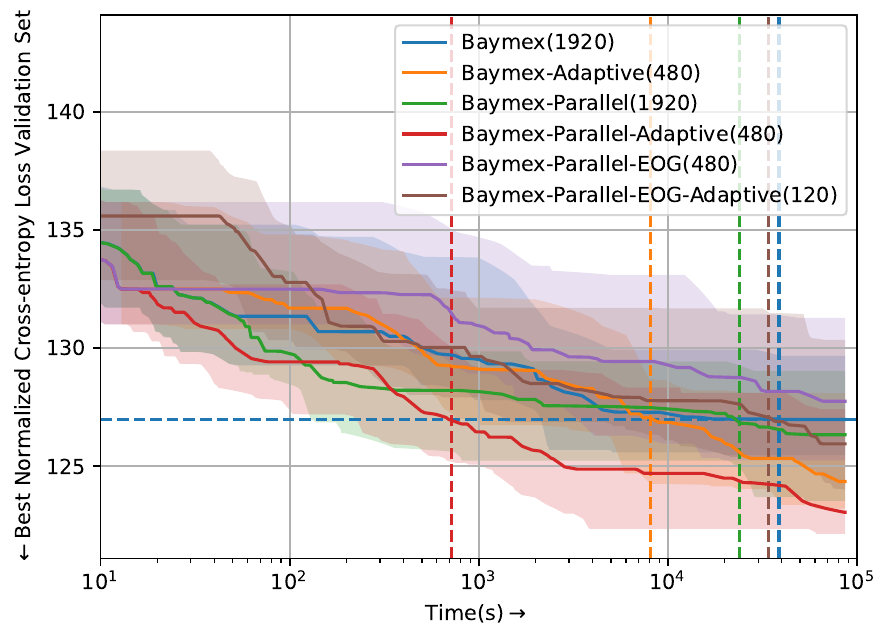}
    \vspace*{-5mm}
    \caption{Convergence of best cross-validation value in the validation elitist
    archive for serial and parallel (EOG/adaptive) Baymex on SUPPORT.
    Solid lines: median. Shaded regions: 25th-75th quantiles. Dashed
    lines: time to reach original serial Baymex's lowest loss.}
    \Description{ }
    \label{fig:adaptive_steering_parallel}
    \vspace*{-3.5mm}
\end{figure}

Beyond speeding up evaluations, an important question is how the enhancements
affect accuracy over time, as they also modify algorithmic behavior.
Figure~\ref{fig:adaptive_steering_parallel} shows the normalized cross‑entropy
loss of the best solution in the validation elitist archive over time for the different Baymex variants. For each variant, we present the configuration with IMS minimum population size for which the median (taken over 30 runs) training accuracy of the best‑in‑front solution found by serial Baymex is matched in the shortest median time.

The three best performing variants all use adaptive steering.
Over time, the adaptive variants increasingly outperform their non-adaptive counterparts.
Baymex-Parallel with adaptive steering attains the best median validation loss within 24 hours.

Baymex with adaptive steering reaches validation cross-entropy values comparable to those of serial Baymex approximately 1.6 times faster.
Combined with parallel execution and EOG updates, the acceleration is up to $\approx 54\times$. 
To assess whether these efficiency gains also translate into better performance at different wall‑clock budgets, we perform statistical tests at the end of the 24‑hour budget as well as after 1 hour and after 5 minutes.
Table~\ref{tab:adaptive_steering_final_loss} confirms that parallelized Baymex with adaptive steering achieves better cross‑validation losses on all wall-budgets inspected. 
Furthermore, parallelism yields statistically significantly faster convergence, as the best
result after 5 minutes is obtained by parallel Baymex. 
For larger datasets this difference is expected to only increase.

\begin{table}[htbp!]
\vspace*{-2mm}
\centering
\tabcolsep=0.6mm
\begin{tabular}{|l|c|c|c|}
\hline
 \textbf{Baymex variant} & \textbf{5 Min} & \textbf{1 Hour} & \textbf{24 Hours} \\ \hline
{\small Baymex}                        & {\small 131(129-135)}          & {\small 128(126-130)}          & {\small 127(125-130)} \\ \hline
{\small Baymex-Adaptive}               & {\small 131(129-134)}          & {\small 128(126-131)}          & {\small 124(123-128)} \\ \hline
{\small Baymex-Parallel}               & {\small \textbf{128}(126-130)} & {\small 128(125-129)}          & {\small 126(124-129)} \\ \hline
{\small Baymex-Parallel-Adaptive}      & {\small \textbf{129}(126-132)} & {\small \textbf{125}(123-129)} & {\small \textbf{123}(122-127)} \\ \hline
{\small Baymex-Parallel-EOG}           & {\small 132(131-136)}          & {\small 130(127-133)}          & {\small 128(125-131)} \\ \hline
{\small Baymex-Parallel-EOG-Adaptive}  & {\small 130(129-133)}          & {\small 128(127-132)}          & {\small 126(124-130)} \\ \hline
\end{tabular}
\caption{The median validation cross-entropy loss and (25th-75th quantiles in brackets) at different wall-clock times. Values are rounded for readability. Bold entries indicate the statistically best performing methods.}
\label{tab:adaptive_steering_final_loss}
    \vspace*{-8mm}
\end{table}

\subsection{Clinical Data Classification}
For the final clinical data classification tasks, we used Baymex-Parallel with adaptive steering with an IMS minimum population size of 480 as the preferred variant of Baymex. Below, when we refer to Baymex, we mean this variant.

\subsubsection{General Baymex Observations}
\begin{figure*}[htbp!]
    \centering
    \begin{subfigure}[t]{0.32\textwidth}
        \centering
        \includegraphics[width=\linewidth]{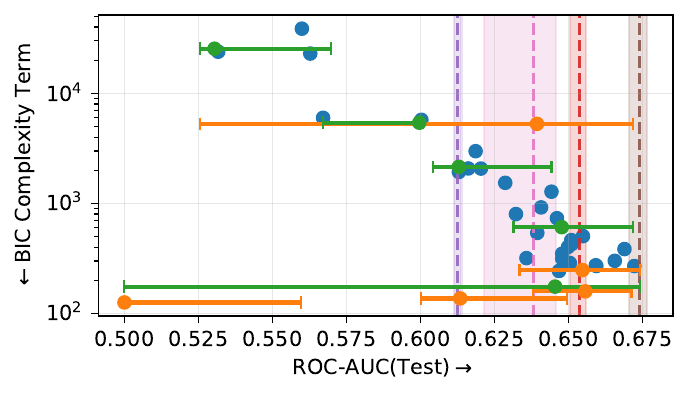}
    \end{subfigure}
    \begin{subfigure}[t]{0.32\textwidth}
        \centering
        \includegraphics[width=\linewidth]{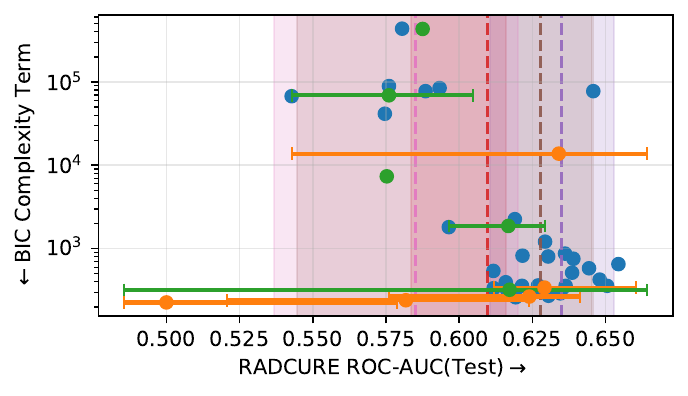}
    \end{subfigure}
    \begin{subfigure}[t]{0.32\textwidth}
        \centering
        \includegraphics[width=\linewidth]{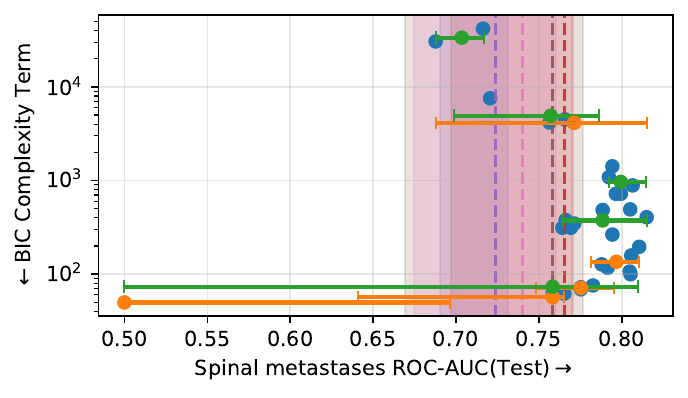}
    \end{subfigure}

    \vspace{-2.5mm}

    \begin{subfigure}[t]{0.5\textwidth}
        \centering
        \includegraphics[width=\linewidth]{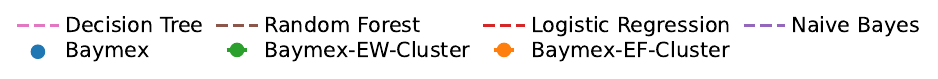}
    \end{subfigure}

    \vspace{-6mm}

    \caption{The min-max range and median of the test scores for the complexity-based clusters, shown in comparison to the models selected based on the best validation/refinement score. Left: SUPPORT, middle:RADCURE, right:Spinal metastases.}
    \label{fig:cluster_distribution}
    \Description{ }
    \vspace*{-2mm}
\end{figure*}

Results for all tested algorithms are summarized in Table~\ref{tab:all_roc_auc_test}. For Baymex, both the
refinement approach and the validation approach are shown. As hypothesized, for larger datasets, the refinement
approach is (slightly) preferable, whereas the converse is true for smaller datasets. In
Figure~\ref{fig:cluster_distribution}, additional details are shown for all algorithms, with in particular
extra results for Baymex showcasing the performance along the front via grouping models in separate complexity
clusters. Here, only the most appropriate strategy is shown (i.e., refinement for SUPPORT, validation for
RADCURE and spinal metastases).

\subsubsection{SUPPORT}
\label{ss:results_support}
\begin{figure}[htbp!]
    \vspace*{-2mm}
    \centering
    \includegraphics[width=1.0\linewidth]{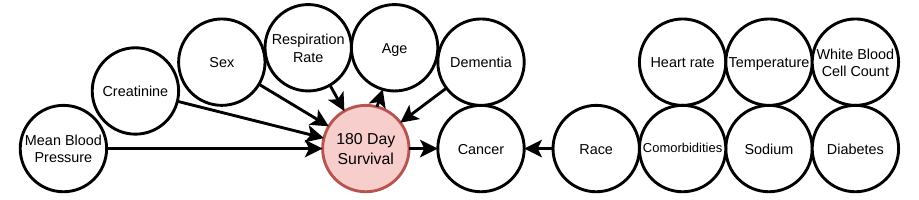}
    \vspace*{-8mm}
    \caption{The best found BN classifier structure for SUPPORT (ROC-AUC Test=0.67). Only features connected to the event node are used by Baymex to predict the outcome.}
    \Description{ }
    \label{fig:support_bn_structure}
    \vspace*{-4mm}
\end{figure}

We ran Baymex for each split for 24 hours.
The BN classifier with the best ROC-AUC Test score across 30 splits is shown in Figure~\ref{fig:support_bn_structure}. Baymex identified an arc between Race and Cancer, which is plausible given that race often correlates with cancer type, see, e.g.,~\cite{thomas2025genomic}. Importantly, we note that the BN captures statistical dependencies that improve predictive performance, not necessarily causal relationships.

Table~\ref{tab:all_roc_auc_test} shows that RF statistically outperforms all competing methods, although the difference with LR and Baymex is extremely small.
When excluding RF due to its limited interpretability, Baymex is statistically tied with LR.

RF shows a marginally higher ROC‑AUC for SUPPORT. While Baymex has a powerful optimization engine, this result is not entirely unexpected: ensemble trees excel on heterogeneous tabular datasets and are less affected by structural uncertainty. Baymex, by contrast, must infer a probabilistic dependency structure, making it more sensitive to limited signal or noisy features. This can result in RF holding a slight edge despite overall comparable performance. 

\begin{table}[t]
\centering
\begin{tabular}{|l|c|c|c|}
\hline
\small{\textbf{Algorithm}}  & \small{\textbf{SUPPORT}} & \small{\textbf{RADCURE}} & \small{\textbf{Spinal}} \\ \hline 
\small{DT}                 & \small{0.64(0.62-0.65)}          & \small{0.59(0.57-0.60)}          & \small{0.74(0.73-0.74)}          \\ \hline
\small{RF}                 & \small{\textbf{0.67(0.67-0.68)}} & \small{\textbf{0.63(0.61-0.63)}} & \small{0.76(0.74-0.77)}          \\ \hline
\small{LR}                 & \small{0.65(0.65-0.66)}*         & \small{0.61(0.60-0.61)}          & \small{\textbf{0.77(0.75-0.77)}} \\ \hline
\small{NB}                 & \small{0.61(0.61-0.61)}          & \small{\textbf{0.64(0.63-0.64)}} & \small{0.72(0.72-0.72)}          \\ \hline
\begin{tabular}[c]{@{}l@{}}\small{Baymex}\\ \small{(Refinement)} \end{tabular}
                   & \small{0.65(0.65-0.65)}*         & \small{0.60(0.59-0.62)}          & \small{\textbf{0.77(0.74-0.78)}} \\ \hline
\begin{tabular}[c]{@{}l@{}}\small{Baymex}\\ \small{(Validation)} \end{tabular} 
                   & \small{0.65(0.62-0.65)}          & \small{\textbf{0.62(0.61-0.64)}} & \small{\textbf{0.79(0.77-0.80)}} \\ \hline
\end{tabular}
\caption{The median ROC-AUC of the SUPPORT, RADCURE, and spinal metastases test sets. Numbers between parenthesis are the 25th-75th quantiles. Bold entries indicate the statistically best performing methods. An asterisk (*) indicates the statistically best performing method on the SUPPORT dataset when RF is excluded.}
\label{tab:all_roc_auc_test}
    \vspace*{-8mm}
\end{table}

\subsubsection{RADCURE}
\label{ss:results_radcure}
\begin{figure}[htbp!]
    \centering
    \includegraphics[width=1.0\linewidth]{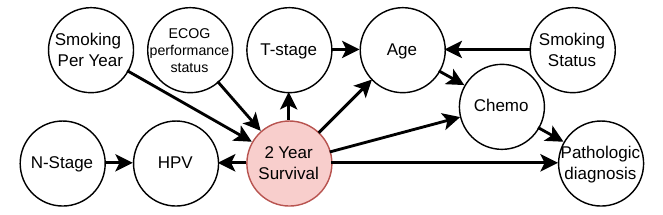}
    \vspace*{-8mm}
    \caption{The best found BN classifier structure for RADCURE (ROC-AUC(Test)=0.68). Only features connected to the event node are used by Baymex to predict the outcome.}
    \Description{ }
    \label{fig:radcure_bn_structure}
    \vspace*{-4mm}
\end{figure}

We ran Baymex for each split for 24 hours. 
The BN classifier with the best ROC-AUC test score across 30 splits is shown in Figure~\ref{fig:radcure_bn_structure}.
In addition to the direct predictors of 2‑year survival (HPV, Smoking per Year, ECOG performance status, T-stage, Age, Chemo, and Pathologic diagnosis), Baymex also identified several dependencies among clinical variables. These include arcs between N-stage and HPV, T-stage and Age, Smoking Status and Age, Age and Chemo, and Chemo and Pathologic diagnosis. 
Except for Pathologic diagnosis, all of these relationships are clinically plausible and consistent with known patterns in head‑and‑neck oncology~\cite{effect_hpv_on_survival_1, effect_age_and_chemo_on_survival, effect_tnm_and_age_and_smoking_on_survival, effect_ecog_on_survival_1, effect_ecog_on_survival_2}.

Table~\ref{tab:all_roc_auc_test} shows that, Baymex has the best performance among all tested algorithms, statistically tied with RF and NB. Moreover, in Figure~\ref{fig:cluster_distribution}, we see that multiple clusters contain solutions with even better test ROC-AUC scores.

\subsubsection{Spinal Metastases}
\label{ss:results_spinal}
\begin{figure}[htbp!]
    \vspace*{-4mm}
    \centering
    \includegraphics[width=1.0\linewidth]{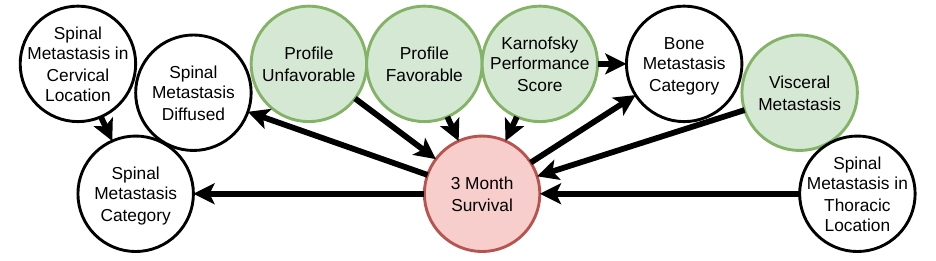}
    \vspace*{-8mm}
    \caption{The best found BN classifier structure for spinal metastases (ROC-AUC Test=0.83).
    Features in green correspond to the four most important features identified in~\cite{dataset_spinal}.}
    \Description{ }
    \label{fig:spinal_bn_structure}
    \vspace*{-4mm}
\end{figure}

We ran Baymex for each split for 6 hours. The BN classifier with the best ROC-AUC Test score across 30 splits is shown in Figure~\ref{fig:spinal_bn_structure}. A comparison with literature reveals agreement with previously determined clinically relevant factors~\cite{dataset_spinal}.
Specifically, Baymex identifies all four features reported as most important, namely: the Karnofsky score, the presence of visceral metastases and clinical profile variables.

Table~\ref{tab:all_roc_auc_test} shows that Baymex performs best among all algorithms, statistically tied with LR. Also, Figure~\ref{fig:cluster_distribution} shows again multiple clusters containing solutions with even better test ROC-AUC scores.

\section{Limitations and Future Work}
\label{s:discussion}
While parallelization and adaptive steering proved effective in improving
Baymex’s efficiency, additional gains may be achieved by exploiting a key
strength of its internal search engine, GOMEA: partial evaluations. In earlier
work where BNs were optimized without a classification task, partial evaluations were not implemented because one of the optimized
objectives (the Kullback–Leibler Divergence~\cite{kl_divergence_book}) lacks the
decomposability required for incremental score updates. In contrast, BN
classification with cross‑entropy loss and BIC complexity is decomposable into
local score contributions for both objectives. However, preliminary experiments
indicate that partial evaluations may introduce a new
bottleneck: substantially increased memory consumption due to the need to retain
discretized data and associated probability tables per solution. Addressing this
challenge remains an important engineering task.

Beyond computational efficiency, this study demonstrates that Baymex can learn
competitive BN classifiers from real‑world clinical data, but it does not yet
leverage Baymex’s multi‑objective capabilities for human‑centered model
selection or knowledge discovery. A natural next step is to integrate domain
experts into the evaluation loop to assess whether the learned models align with
expert knowledge or meaningfully influence expert beliefs.

Finally, future work should explore how Baymex can support clinical practice
more directly by enabling transparent model selection and
facilitating joint knowledge discovery with clinicians.

\section{Conclusion}
\label{s:conclusion}
In this work, we enhanced Baymex, an evolutionary multi‑objective algorithm for Bayesian network structure learning and discretization, and adapted it for BN classification. We showed that parallelization and adaptive steering can substantially accelerate the search procedure. Across three real‑world clinical prediction tasks, including our in‑house spinal metastases case, Baymex achieved performance that is competitive with, and often superior to, established machine learning methods such as decision trees, logistic regression, naïve Bayes, and random forests. Importantly, Baymex also yields transparent models whose learned dependencies can provide clinically meaningful insights, supporting explainable AI while remaining state of the art for clinical classification.

\begin{acks}
This research is part of the research programme Open Competition Domain
Science-KLEIN (project number OCENW.KLEIN.111), which is financed by the
Dutch Research Council (NWO). Further, we thank SURF (www.surf.nl) for the support in using the National Supercomputer Snellius. T. Schlender was supported by the Gieskes-Strijbis Fonds (project "Uitlegbare Kunstmatige Intelligentie").
\end{acks}

\bibliographystyle{ACM-Reference-Format}
\bibliography{bib,bib_literature}

\end{document}